\documentclass[preprint,12pt]{elsarticle}

\usepackage[colorlinks,
            linkcolor=red,
            anchorcolor=blue,
            citecolor=green,
           backref=page
            ]{hyperref}
\usepackage{amssymb}
\usepackage{amsmath}

\usepackage{multirow}

\usepackage[linesnumbered,ruled]{algorithm2e}
\usepackage{amssymb}

\usepackage{diagbox}
\journal{EAAI}

\begin{document}

\begin{frontmatter}



\title{Purifying Naturalistic Images through a Real-time Style Transfer Semantics Network}


\author[label1]{Tongtong Zhao}
\ead{zhaotongtong@dlmu.edu.cn}
\author[label1]{YuXiao Yan}
\ead{yuxiaoyan@dlmu.edu.cn}
\author[label1]{Ibrahim Shehi Shehu }
\ead{shehu@dlmu.edu.cn}
\author[label1]{Xianping Fu\corref{cor1}}
\ead{fxp@dlmu.edu.cn}
\author[label1]{Huibing Wang\corref{cor1}}
\ead{huibing.wang@dlmu.edu.cn}
\cortext[cor1]{Both Xianping Fu and Huibing Wang are corresponding authors.}

\address[label1]{Information Science and Technology College, Dalian Maritime University, \\ Dalian 116026, China}
\address{}

\begin{abstract}

Recently, the progress of learning-by-synthesis has proposed a training model for synthetic images, which can effectively reduce the cost of human and material resources. However, due to the different distribution of synthetic images compared to real images, the desired performance cannot still be achieved. Real images consist of multiple forms of light orientation, while synthetic images consist of a uniform light orientation. These features are considered to be characteristic of outdoor and indoor scenes, respectively. To solve this problem, the previous method learned a model to improve the realism of the synthetic image. Different from the previous methods, this paper takes the first step to purify real images. Through the style transfer task, the distribution of outdoor real images is converted into indoor synthetic images, thereby reducing the influence of light. Therefore, this paper proposes a real-time style transfer network that preserves image content information (eg, gaze direction, pupil center position) of an input image (real image) while inferring style information (eg, image color structure, semantic features) of style image (synthetic image). In addition, the network accelerates the convergence speed of the model and adapts to multi-scale images. Experiments were performed using mixed studies (qualitative and quantitative) methods to demonstrate the possibility of purifying real images in complex directions. Qualitatively, it compares the proposed method with the available methods in a series of indoor and outdoor scenarios of the LPW dataset. In quantitative terms, it evaluates the purified image by training a gaze estimation model on the cross data set. The results show a significant improvement over the baseline method compared to the raw real image.

\end{abstract}


\end{frontmatter}


\section{Introduction}\label{sec:introduction}

The accuracy of gaze estimation under indoor conditions is currently about 0.5 to 1 degree, which is commendable, but the accuracy of gaze estimation under outdoor conditions is still unsatisfactory, between 8 and 10 degrees. However, appearance-based gaze estimation has recently progressed under outdoor conditions by using a large-scale real-image training data set with annotations through recent rises in high-capacity deep convolution networks. In addition, annotating training data sets requires a lot of manual labor. To solve this problem, a training model on a synthetic image is preferred because the annotations are automatically available. There are four main types of eye image synthesis methods: optical flow, three-dimensional eye reconstruction, eye model method, and GAN (Generation Against Network). But this solution has a drawback, the distribution between the real image and the synthetic image is quite different. The distribution of synthetic images is more prone to indoor lighting, with slight variations depending on the synthesis method. On the other hand, due to the interference of light and other external factors, the distribution of real images is more complicated (prone to outdoor lighting), making the distribution of real images difficult to learn. Therefore, using synthetic images for training, the effects of testing in real scenes or on real image data sets will not be satisfactory. One solution is to attenuate the distribution of real images by improving the simulator, which can be expensive and time consuming. Another solution is to use unmarked actual data to improve the authenticity of the synthetic image from the simulator. This method cannot be applied to outdoor (field) scenes due to its weak training time and adaptability to different situations in the field.

In a different manner, we see the image distribution change between the real image and the synthetic image as a gap from the previous solution. The distribution of synthetic images is more regular and easy to learn, rather than trying to improve the realism of the synthetic image, we prefer to purify the real image to make it similar to the indoor scene, while retaining annotation information such as the gaze direction. Considering the distribution of images as its style information and annotation information as its content information, the problem can be considered as a style transfer task. Based on this, we propose a controllable neural transfer architecture to purify real images.

Similar to Gatys et al.\cite{Gatys2015}, they proposed a new method of using neural networks to capture artistic image styles and transfer them to real-world photos, and Feifei Li et al.\cite{lifeifei2016}, who proposed using a perceptual loss function to train feed-forward networks for image transformation tasks. Our approach not only uses advanced feature representations of images from the hidden layers of the VGG convolutional network to separate and recombine content and style, but also trains the feed-forward network to better calculate the loss of content and style. The main difference with Gatys et al.\cite{Gatys2015}  and Feifei Li et al.\cite{lifeifei2016} is that images for gaze prediction need more-precise content information and more emphasis on image spatial arrangement of reservations, as such we consider the image distribution variation between real and synthetic images as the gap between the indoor and the outdoor situation and propose a real-time style transfer with semantics network for purifying real images.

Our image purification architecture can be divided into three parts: coarse segmentation network, feature extraction network and loss network. We train the coarse segmentation network to segment the pupil and iris regions to avoid "orphan semantic tags" that only appear in the input image. Due to the outdoor lighting effect, the "orphan tag" is usually the pupil area, and we limit the pupil semantic area to the center of the iris area. We also observed that segmentation does not require pixel precision because the final output is constrained by our loss network.

For feature extraction, our work is most directly related to the work initiated by Gates et al.\cite{Gatys2015}. The feature map of the deep convolutional neural network with differentiated training is used to achieve the breakthrough performance of the transfer of painting style. We train a feed-forward feature extraction network for image transformation tasks. However, we did not directly use the network of Feifei Li et al.\cite{lifeifei2016}, but modified the network to delete the checkerboard. Our image conversion task network consists of four residual blocks. All non-residual convolutional layers are followed by bulk normalization and ReLU non-linearity, except for the output layer, which uses scaled tanh to ensure that the output has pixels in the range [0,255]. The first and last layers use $9\times9$ cores; all other deconvolution layers use $4\times4$ cores to fill 1 and the convolution layer uses $3\times3$ kernels to fill 0. The main difference with the network of Feifei Li et al.\cite{lifeifei2016} is our network that aims to learn as much as possible on the premise of synthetic distribution, to minimize the loss of content transmission, and to solve the problem of insufficient spatial alignment information caused by the gram matrix. To achieve this goal, we propose a loss network with a novel loss function that makes some key modifications to the standard perceptual loss to maximize the content of the real image and the distribution of the synthetic image. Our network not only considers red, green, and blue (RGB) color channels, but uses color and semantic representations for style transformation. Through the semantic features, we can solve the spatial arrangement information and avoid the spatial configuration that the image is destroyed due to the style transformation.

Our contributions are presented in this paper in three folds:

1. We took the first step to consider the image distribution variation between real and synthetic images as the gap between the indoor and outdoor situation and propose image purification network architecture to purify the real image, making it similar to indoor conditions while retaining annotation information.

2. We proposed a loss network with a novel loss function, with some key modifications to the standard perceptual loss to maximize the content of the real image and the distribution of the synthetic image. Our network not only considers the RGB color channel, but uses the representation of color and semantic features for style conversion. Through the semantic features, we can solve the spatial arrangement information and avoid the spatial configuration that the image is destroyed due to the style conversion.

3. We proposed a hybrid research method (qualitative and quantitative) for experiments. The results show that the proposed architecture significantly purifies the real image compared with the existing methods. We used different gaze estimation methods to achieve improved results on cross-data sets.

\section{Related Works}\label{sec:relatedworks}

In general, there are two main types of eye gaze estimation methods: feature-based and appearance-based\cite{hansen2010in}. Feature-based methods are intended to identify local features of the eye, such as contours, corners of the eye, and reflections from images of the eye. Pupil and corneal reflexes are commonly used for eye localization. Calibration with high-resolution cameras and other specialized hardware such as synchronized cameras and light sources can extract more precise geometric properties. However, the gaze feature is not stable enough under natural light.

\subsection{Appearance-based gaze estimation}

The appearance-based approach is believed to work better under natural light. These methods use image content as input to map these directly to screen coordinates. A regression model of the gaze can then be constructed, by which we can obtain the position of the gaze or the two-dimensional rotation angle when using the new eye image as input. Compared to feature-based methods, the appearance-based approach does not require any dedicated hardware and exhibits good robustness to outliers. Most appearance-based methods extract feature vectors of cropped eye images and map them into low-dimensional spaces. Then the regression model in this space can be constructed.

Recent studies aim to better represent the appearance, and Lu et al. \cite{lu2014adaptive} proposed a low-dimensional feature extraction method. It divides the eye area into three columns and five columns and calculates the gray value and the percentage of each area. Therefore, a 15-dimensional feature vector is defined. However, this feature does not apply to eye images under free head movement. Wang et al.\cite{wang2016appearance} introduced a deep feature extracted from convolutional neural networks. The deep feature has sparse characters and provides a effective solution for gaze estimation.

\subsection{Eye image synthesis}
There are four main categories of eye image synthesis methods: Optical Flow\cite{lu2015gaze}\cite{Wang2014hierarchical}, 3D eye reconstruction\cite{Sugano2014learning}\cite{Wood2015rendering}, Model-based method \cite{Wood2016a3d} and GANs (Generative Adversarial Networks)\cite{Shrivastava2016learning}.

The eye image synthesis process in Lu et al.\cite{lu2015gaze} used 1D flows to simulate the appearance distortion caused by head pose moving, and Wang et al.\cite{Wang2014hierarchical} introduced a 2D interpolation to synthesize the eye appearance variation caused by eyeball moving. These optical flow methods treat eye image synthesis as optical shift of original image and could not be utilized under large head rotation. Generating eye images by 3D eye reconstruction is highly depending on the pre-trained face 3D model. Sugano et al.\cite{Sugano2014learning} recovered multi-view eye images from 3D shapes of eye region reconstructed from 8 cameras eye image capture system. While Wood et al.\cite{Wood2015rendering} relied on high-quality head scans to collect high resolution eye images. In order to generate multi-part eye images, Wood et al.\cite{Wood2016a3d} also presented a morphable model of the facial eye region, as well as an anatomy-based eyeball model. Model-based method tunes parameters to obtain high resolution eye images, which are coincide with the ground truth situation. Shrivastava et al.\cite{Shrivastava2016learning} used GANs to generate synthetic eye images using unlabeled real data and learnt a refiner model that improves the realism of these synthetic images. While GANs output different synthetic image by same input image, it is still not controllable to generate image with specific gaze angle.

\subsection{Learning-by-synthesis}
Learning-based methods perform well in appearance-based gaze estimation but require large amounts of training data. Learning-by-synthesis approaches were proposed to solve this  problem. Wood et al.\cite{Wood2016learning} presented a novel method to synthesize large amounts of variable eye region images as training data, which addressed the limitation of learning-by-synthesis with respect to the appearance variability and the head pose and gaze angle distribution. Other works learn a feature representation in feature space. For instance, Wang et al.\cite{Wang2014hierarchical} proposed an appearance-based gaze estimation method by supervised adaptive feature extraction\cite{AIterative}\cite{ALearning}\cite{AMultiview}\cite{ARobust}\cite{ASemantic} and hierarchical mapping model\cite{ACycle-Consistent}\cite{ADeep}\cite{AEffective}, during which appearance synthesis method is proposed to increase the sample density. Zhang et al.\cite{wild2017} introduced a CNN-based gaze estimation method, which concatenated head pose vector in the hidden layer of neural network. This change improved the performance of CNN-based gaze estimation training by synthetic image dataset. Sugano et al.\cite{Sugano2014learning} presented a learning-by-synthesis approach for appearance-based gaze estimation and trained a 3D gaze estimator by a large amount of cross-subject training data. In their experiments, $k$-nearest neighbor was selected as comparison, from which we can see that $k$-NN regression estimators can perform well with a large amount of dense training samples.

\subsection{Style Transfer}

Previous methods learn a model to improve the realism of synthetic images, instead we take the first step to purify real images to weaken the influence of light and convert the distribution of outdoor real image to that of indoor synthetic image. This can be seen as a style transfer task, global style transfer algorithms process an image by applying a spatially-invariant transfer function. Reinhard et al.\cite{Reinhard2001} match the means and standard deviations between the input and reference style image after converting them into a decorrelated color space. Local style transfer algorithms based on spatial color mappings are more expressive and can handle a broad class of applications such as transfer of artistic edits\cite{Gatys2015}\cite{Selim2016}, weather and season change\cite{Gardner2015}. Similar to Gatys et al.\cite{Gatys2015}, which proposed a novel approach using neural networks to capture the style of artistic images and transfer it to real-world photographs, and Feifei Li et al.\cite{lifeifei2016} which proposed the use of perceptual loss function for training feed-forward networks for image transformation tasks. Our approach not only uses high-level feature representations of images from hidden layers of the VGG convolutional network to separate and reassemble content and style but also trains feed-forward networks to better calculate the loss of content and style.

\section{Proposed Method}\label{sec:Proposedmethod}

We briefly reviewed the style transfer approach introduced by Feifei Li et al.\cite{lifeifei2016} that transfers the reference style image S into the input image I and then generate a stylized image O by minimizing the perceptual losses. Meanwhile, we briefly review the style transfer approach introduced by Gatys et al.\cite{Gatys2015} that transfers the reference style image S into the input image I and then generate a stylized image O by minimizing the objective function consisting of a content loss and a style loss.
\begin{figure*}[!htbp]
\centering
 \begin{minipage}[]{1\textwidth}
    \centering
     \includegraphics[width = 1\textwidth,angle=0]{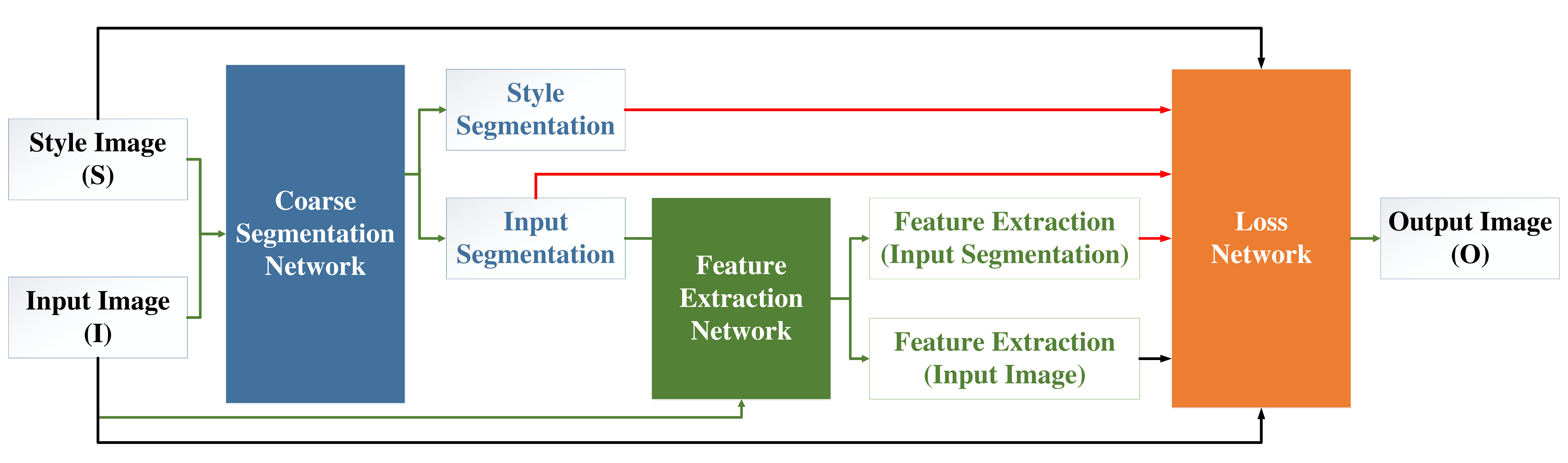}
 \end{minipage}
    \caption{The overview of proposed method. Our method can be divided into three parts: coarse segmentation network, feature extraction network and loss network. }
    \label{figoverview}
\end{figure*}

The key idea of  Feifei Li et al.\cite{lifeifei2016} is that method consists of two components: an image transformation network $f_{W}$ and a loss network $\phi$, $f_{W}$ is a deep residual convolutional network with weights $W$, it transforms I into O via mapping $O=f_{W}(I)$. Loss network $\phi$ is used to minimize the loss between O and I, O and S with perceptual loss method. Loss between O and I is denoted as  feature reconstruction loss $\ell_{feat}$ which can be represent as :
\begin{equation}
\ell_{feat}^{\phi,j}(O,I)=\frac{1}{C_{j}H_{j}W{j}}\|\phi_{j}(O)-\phi_{j}(I)\|_{2}^{2}
\end{equation}
where j is a convolutional layer and  $\phi_{j}(\cdot)$ is a feature map of shape $C_{j}\times H_{j}\times W_{j}$. Loss between O and S is denoted as style reconstruction loss which is the squares Frobenius norm of the difference between the Gram matrices(similar with \cite{Gatys2015}) of O and I:
\begin{equation}
\ell_{style}^{\phi,j}(O,S)=\|G_{j}^{\phi}(O)-G_{j}^{\phi}(S)\|_{2}^{2}
\end{equation}
 The Gram matrix can be computed efficiently by reshaping $\phi_{j}(\cdot)$ into a matrix $\psi$ of shape $C_{j}\times H_{j}\times W_{j}$; then $G_{j}^{\phi}(\cdot)=\frac{\psi\psi^{T}}{C_{j}\times H_{j}\times W_{j}}$. Image O is generated by solving the problem
 \begin{equation}
O=\arg\min_{I}\alpha\ell_{feat}^{\phi,j}(O,I)+\beta\ell_{style}^{\phi,j}(O,S)+\theta\ell_{TV}(I)
\end{equation}
where $\alpha$,$\beta$,$\theta$ are scalars, I is initialized with white noise, and optimization is performed using L-BFGS.

The key idea of Gatys et al. \cite{Gatys2015} is that features extracted by a convolutional network carries information about the content of the image, while the correlations of these features encode the style. The Objective function can be represented as:
\begin{equation}
L_{total}=\sum_{l=1}^L\alpha_lL_{content}^l+\sum_{l=1}^L\beta_lL_{style}^l
\end{equation}
where L is the total number of convolutional layers and $l$ indicates the $l$-th convolutional layer of the deep convolutional neural network. $\alpha_{l}$ and $\beta_{l}$ are the weights to configure layer preferences. Each layer with $N_{l}$ distinct filters has $N_{l}$ feature maps each of size $M_{l}$, where $M_{l}$ is the height times the width of the feature map. So the responses in each layer $l$ can be stored in a matrix
 $F[\cdot] \in R^{N_{l}\times M_{l}}$ where $F[\cdot]_{ij}$ is the activation of the $i^{th}$ filter at position $j$ in each layer $l$. The content loss, denoted as $L_{content}$, is simply the mean squared error between  $F_{l}[O] \in R^{N_{l}\times M_{l}}$ and $F_{l}[I] \in R^{N_{l}\times M_{l}}$ .
 \begin{equation}
L_{content}^l=\frac{1}{N_{l}M_{l}}\sum_{ij}(F_{l}[O]-F_{l}[I])^{2}_{ij}
\end{equation}
The style loss, denoted as $L_{style}$, can be represented as:
 \begin{equation}
L_{style}^l=\frac{1}{N_{l}^{2}}\sum_{ij}(G_{l}[O]-G_{l}[S])^{2}_{ij}
\end{equation}
Gram matrix $G_{l}[\cdot]$ is defined as the inner product between the vectored feature maps which is $F_{l}[\cdot]F_{l}[\cdot]^{T} \in R^{N_{l}\times N_{l}}$.

Our proposed network (As Fig.~\hyperref[figoverview]{1}) takes two images with their mask: an input image which is a real eye image from video of driving environment or real eye image dataset. A stylized and retouched image referred as the reference style image from synthetic image dataset. We use this to train the gaze estimator, as we seek to transfer the style of the reference to the input image while keeping the content and spatial information due to its importance in appearance-based gaze estimation. The proposed architecture can be divided into three parts: coarse segmentation network, feature extraction network and loss network.
\begin{figure*}[!htbp]
\centering
 \begin{minipage}[]{1\textwidth}
    \centering
     \includegraphics[width = 1\textwidth,angle=0]{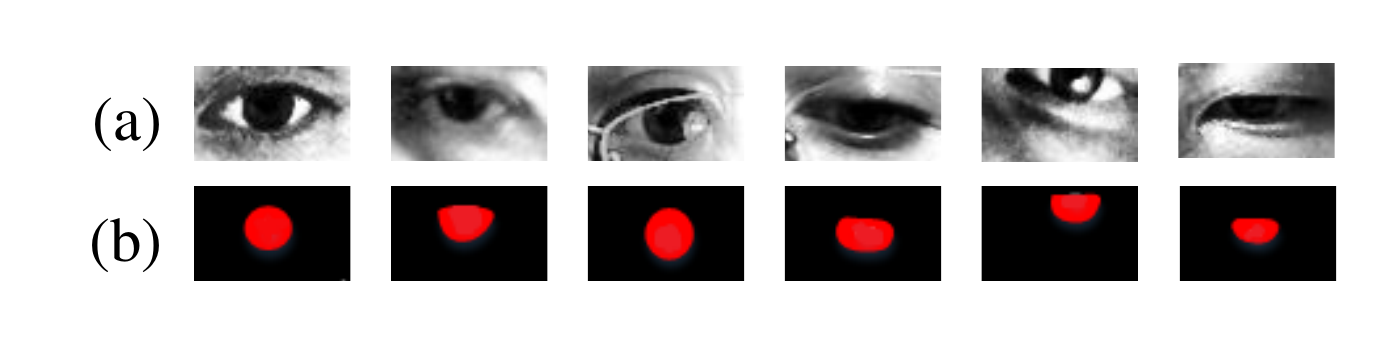}
 \end{minipage}
    \caption{Example of "orphan semantic labels". We can observe that only the iris region (red) is present in the image and the coarse segmentation network could not label the pupil region (white) in this image.}
    \label{figsegmen}
\end{figure*}

\subsection{Coarse segmentation network}

We train fully convolutional networks which according to Long et al.\cite{Long2015} as the coarse segmentation network to segment, according to the line of sight estimate for the human eye image and the consideration of simplifying the task, we only mark two kinds of information on the human eye image: the pupil(white) and the iris(red). However, many human eye images are influenced by light and other factors, and sometimes the pupil and the iris cannot be completely separated, as Fig.~\hyperref[figsegmen]{2} shows that only the iris region is present in the image and the coarse segmentation network could not label the pupil region in this image, thus we called this kind label "orphan semantic labels".

\begin{figure*}[!htbp]
\centering
 \begin{minipage}[]{1\textwidth}
    \centering
     \includegraphics[width = 0.7\textwidth,angle=0]{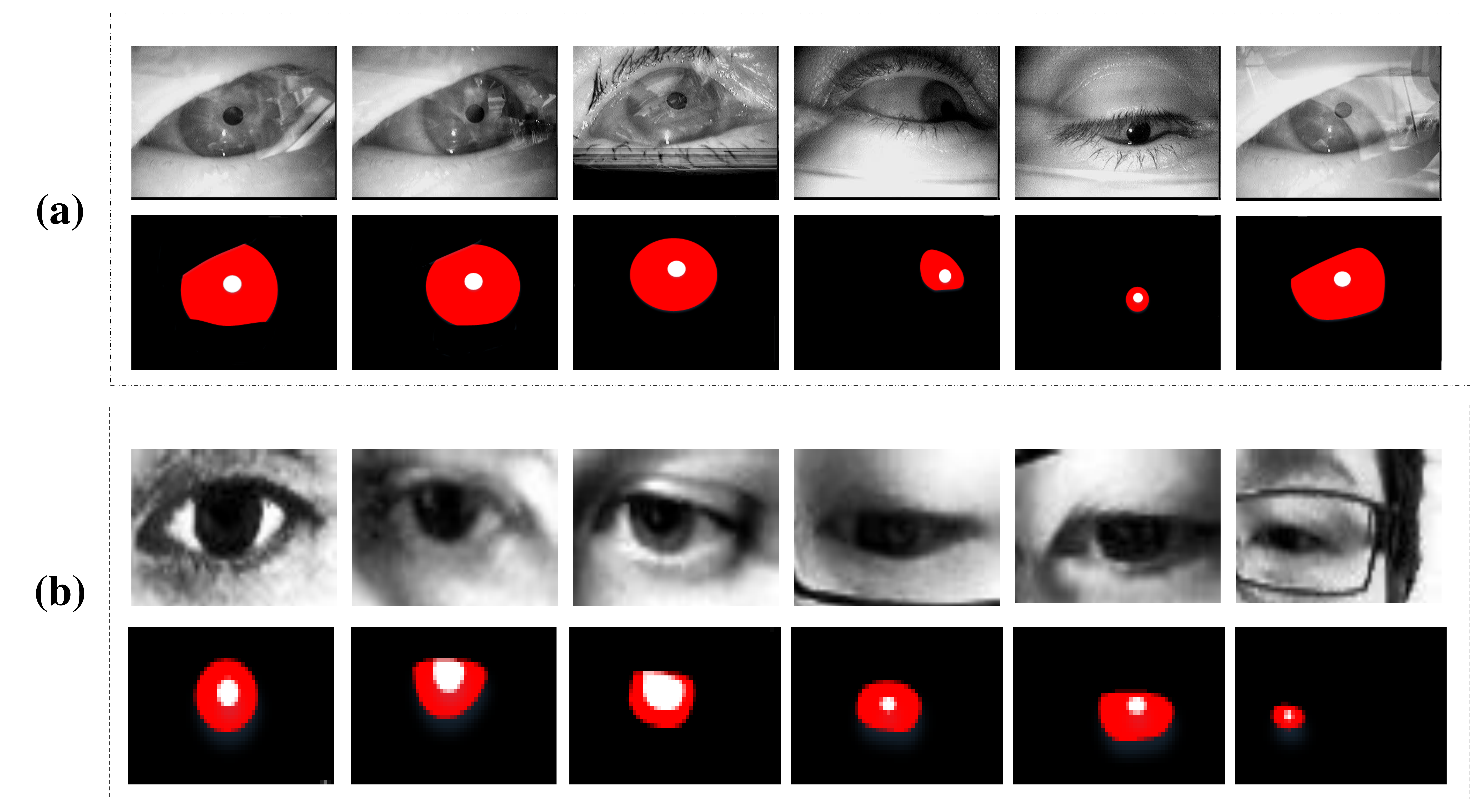}
 \end{minipage}
    \caption{Coarse segmentation on LPW dataset (a) and MPIIGaze dataset(b).Pupil region is labelled on white and iris region is red. We can observe that although the testing dataset under different illumination condition, proposed network can achieve good results without "orphan semantic labels".}
    \label{figsegmen}
\end{figure*}
To avoid "orphan semantic labels", which are caused by outdoor illumination effect on the pupil region, we constrain the pupil semantic region to be set as the center of iris region. The segmentation results without "orphan semantic labels" are shown in Fig 3. We have also observed that the segmentation does not need to be pixel accurate since eventually the output is constrained by our loss network.

\subsection{Feature extraction network}

For feature extraction, our work is inspired by Gatys et al.\cite{Gatys2015} that employs the feature maps of discriminatively trained deep convolutional neural networks to achieve ground breaking performance for painterly style transfer\cite{Selim2016}. The main difference with our work is that, our approach learns as much as possible on the premise of the synthesis distribution, to minimize the loss of content transfer and to solve the problem of the lack of spatial arrangement information caused by the gram matrix.

 \begin{figure*}[!htbp]
\centering
 \begin{minipage}[]{1\textwidth}
    \centering
     \includegraphics[width = 1\textwidth,angle=0]{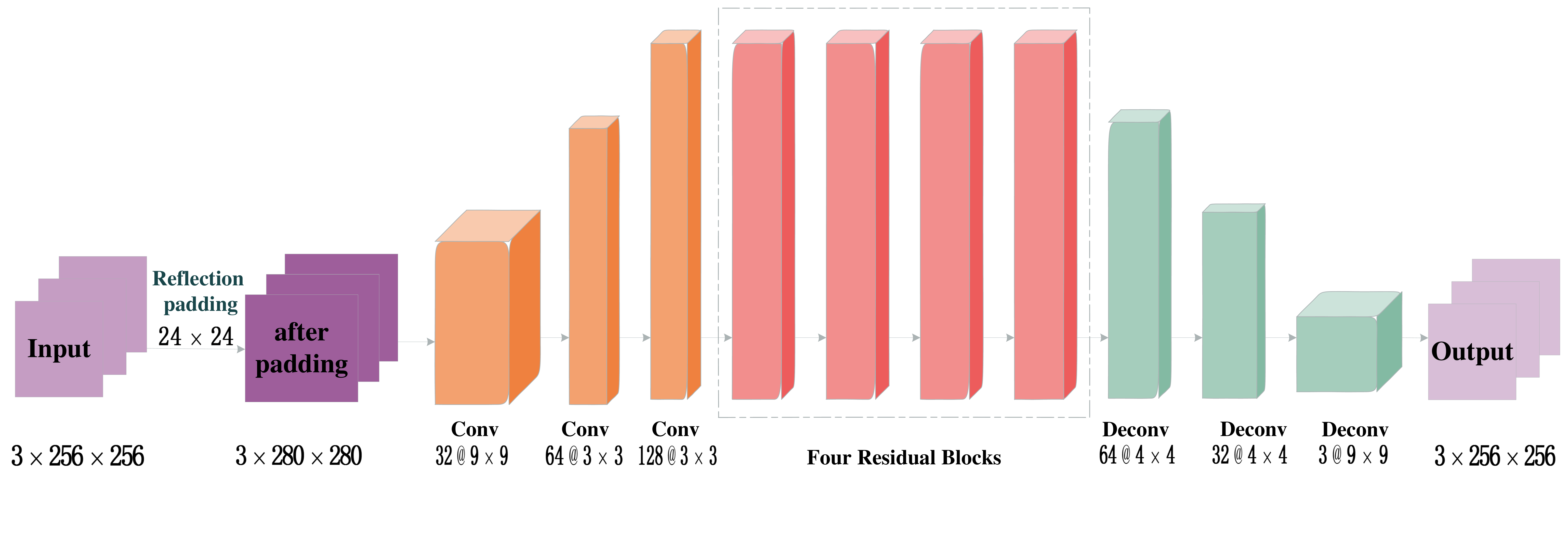}
 \end{minipage}
    \caption{The overview of feature extraction network. The first and last layers use $9\times9$ kernels, all other deconvolutional layers use $4\times4$ kernels with padding 1 and convolutional layers use $3\times3$ kernels with padding 0.}
    \label{figfeature}
\end{figure*}

Similar with Feifei Li et al.\cite{lifeifei2016}, our feature extraction network roughly follow the architectural guidelines set forth by \cite{DCGAN2015}. Feifei Li et al.\cite{lifeifei2016} propose a image transformation network which eschews pooling layers, instead using strided and fractionally strided convolutions for in-network convolution and deconvolution. The image transformation network comprises five residual blocks, all nonresidual convolutional layers are followed by batch normalization and ReLU nonlinearities with the exception of the output layer. However, from \cite{checkboard2016} we know that the standard approach of producing images with deconvolution has some conceptually simple issues that lead to artifacts in produced images. Inspired by \cite{checkboard2016}, we modified the structure of image transformation network\cite{lifeifei2016} to our feature extraction network. The structure can be shown as Fig.~\hyperref[figfeature]{4} and Table 1.

Our network body comprises four residual blocks. All nonresidual convolutional layers are followed by batch normalization and ReLU nonlinearities with the exception of the output layer, which instead uses a scaled tanh to ensure that the output has pixels in the range [0,255].

\begin{figure*}[!htbp]
\centering
 \begin{minipage}[]{1\textwidth}
    \centering
     \includegraphics[width = 0.7\textwidth,angle=0]{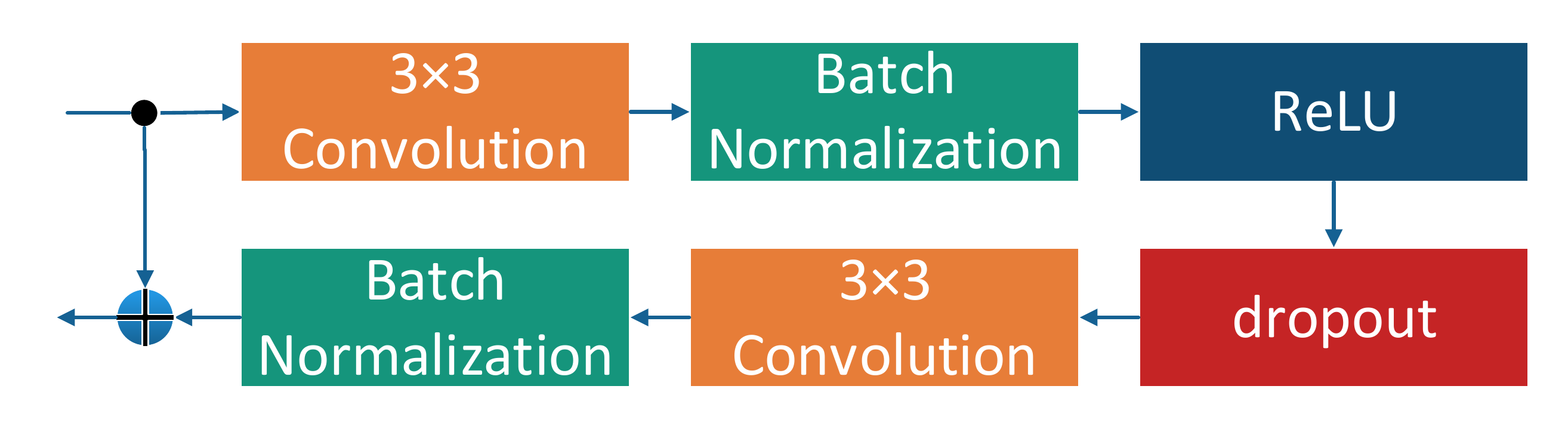}
 \end{minipage}
    \caption{The structure of residual block. Our residual blocks each contain two $3\times3$ convolutional layers with the same number of filters on both layer, similar with \cite{lifeifei2016} but with dropout followed by spatial batch normalization and a ReLU nonlinearity in order to avoid overfitting.}
    \label{figresidual}
 \end{figure*}

The first and last layers use $9\times9$ kernels, all other deconvolutional layers use $4\times4$ kernels with padding 1 and convolutional layers use $3\times3$ kernels with padding 0. We use the residual block design similar with \cite{lifeifei2016} but with dropout followed by spatial batch normalization and a ReLU nonlinearity in order to avoid overfitting, shown in the Fig.~\hyperref[figresidual]{5}.

\subsection{Loss network}

We propose loss network with novel loss function which is a pre-trained VGG-19\cite{Simonyan2014}  network and made some key modifications to the standard perception losses to keep the content of the real images and distribution of the synthetic images to the fullest extent.

\begin{figure*}[!htbp]
\centering
 \begin{minipage}[]{1\textwidth}
    \centering
     \includegraphics[width = 1\textwidth,angle=0]{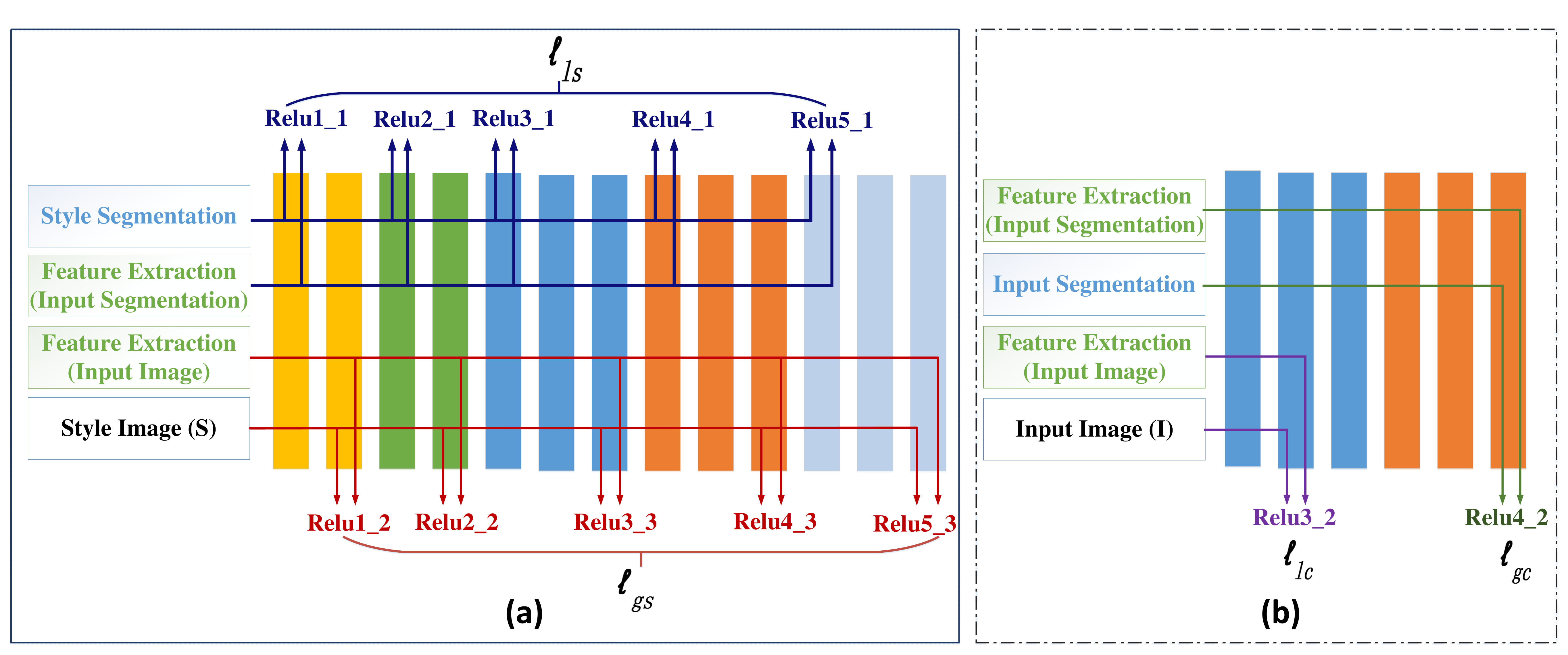}
 \end{minipage}
    \caption{The overview of loss network. In this figure, our loss network can be divided into two parts:  Style reconstruction loss (a) and Feature reconstruction loss(b), feature reconstruction loss is denoted as $\ell_{feat}$ which is the summary of $\ell_{gc}$ and $\ell_{lc}$, meanwhile, style reconstruction loss is denoted as $\ell_{style}$ which is the summary of $\ell_{gs}$ and $\ell_{ls}$. }
    \label{figloss}
\end{figure*}

In Fig.~\hyperref[figloss]{6}, our loss network can be divided into two parts:  Style reconstruction loss (a) and Feature reconstruction loss(b), feature reconstruction loss is denoted as $\ell_{feat}$ which is the summary of $\ell_{gc}$ and $\ell_{lc}$, meanwhile, style reconstruction loss is denoted as $\ell_{style}$ which is the summary of $\ell_{gs}$ and $\ell_{ls}$. As Fig.~\hyperref[figloss]{6} shows that instead of taking only RGB color channels into consideration, our network utilizes the representations of both color and semantic features for style transfer. With the semantic features, we can address the spatial arrangement information and avoid the spatial configuration of the image being disrupted because of the style transformation.
\subsubsection{Feature reconstruction loss}

A limitation of the general content loss is that image structure is not considered when encoding content reconstructions. We address this problem with the image segmentation masks for the input images. With the segmentation mask, the content we prefer to address can be preserved more effectively.  To visualise the image information that is encoded at different layers of the input image with masks, we perform gradient descent on a white noise image to find another image that matches the feature responses of the original image with mask. We then define the squared-error loss between the two feature representations
 \begin{equation}
\ell_{feat}^l=\lambda_{g}\ell_{gc}^l+\lambda_{l}\ell_{lc}^l
\end{equation}
 \begin{equation}
\ell_{gc}^l=\sum_{c=1}^{C}\frac{1}{2N_{l}M_{l}}\sum_{ij}(F_{l}[O]-F_{l}[I])^{2}_{ij}
\end{equation}
 \begin{equation}
\ell_{lc}^l=\sum_{c=1}^{C}\frac{1}{2N_{l}M_{l}}\sum_{ij}(F_{l,c}[O]-F_{l,c}[I])^{2}_{ij}
\end{equation}

where  C is the number of channels in the semantic segmentation mask and $l$ indicates the $l$-th convolutional layer of the deep convolutional neural network, $S_{l,c}[\cdot]$ is the segmentation mask in each layer $l$ with the channel c. $\lambda_{g}$ is the weight to configure layer preferences of global losses $\ell_{gc}$ which calculated between raw input image and features which was extracted by feature extraction network.$\lambda_{l}$ is the weight to configure layer preferences of local losses $\ell_{lc}$ which calculated between input segmentation image and features which was extracted by feature extraction network with the input of segmentation image.

Each layer with $N_{l}$ distinct filters has $N_{l}$ feature maps each of size $M_{l}$, where $M_{l}$ is the height times the width of the feature map. So the responses in each layer $l$ can be stored in a matrix $F[\cdot] \in R^{N_{l}\times M_{l}}$ where $F[\cdot]_{ij}$ is the activation of the $i^{th}$ filter at position $j$ in each layer $l$. The content loss, denoted as $L_{content}$, is simply the mean squared error between  $F_{l}[O] \in R^{N_{l}\times M_{l}}$ and $F_{l}[I] \in R^{N_{l}\times M_{l}}$.
\begin{equation}
F_{l,c}[O]=F_l[O]S_{l,c}[I]
\end{equation}
\begin{equation}
F_{l,c}[I]=F_l[I]S_{l,c}[I]
\end{equation}
As minimizing $\ell_{feat}$, the image content and overall spatial structure are preserved but color, texture, and exact shape are not. Using a feature reconstruction loss for training our image transformation networks encourages the output image O to be perceptually similar to the style image S, without forcing them to match exactly.

\subsubsection{Style reconstruction loss}

Feature Gram matrices are effective at representing texture, because they capture global statistics across the image due to spatial averaging. Since textures are static, averaging over positions is required and makes Gram matrices fully blind to the global arrangement of objects inside the reference image. So if we want to keep the global arrangement of objects, make the gram matrices more controllable to compute over the exact region of entire image, we need to add some texture information to the image. Luan et al.\cite{Deep2017} present a method which add the masks to the input image as additional channels and augment the neural style algorithm by concatenating the segmentation channels, inspired by it, mask is added as the texture information we need to compute over the exact region of entire image, thus the style loss can be denoted as:
\begin{equation}
\ell_{style}^l=\lambda_{g}\ell_{gs}^l+\lambda_{l}\ell_{ls}^l
\end{equation}
 \begin{equation}
\ell_{gs}^l=\sum_{c=1}^{C}\frac{1}{4N^2_{l,c}M^2_{l,c}}\sum_{ij}\left(G_{l}[O]-G_{l}[S]\right)^2_{ij}
\end{equation}
\begin{equation}
\ell_{ls}^l=\sum_{c=1}^{C}\frac{1}{4N^2_{l,c}M^2_{l,c}}\sum_{ij}\left(G_{l,c}[O]-G_{l,c}[S]\right)^2_{ij}
\end{equation}

where C is the number of channels in the semantic segmentation mask and $l$ indicates the $l$-th convolutional layer of the deep convolutional neural network. Each layer with $N_{l}$ distinct filters has $N_{l}$ feature maps each of size $M_{l}$, where $M_{l}$ is the height times the width of the feature map. So the responses in each layer $l$ can be stored in a matrix $F[\cdot] \in R^{N_{l}\times M_{l}}$ where $F[\cdot]_{ij}$ is the activation of the $i^{th}$ filter at position $j$ in each layer $l$.
\begin{equation}
F_{l,c}[O]=F_l[O]S_{l,c}[I]
\end{equation}
\begin{equation}
F_{l,c}[S]=F_l[S]S_{l,c}[S]
\end{equation}
\begin{equation}
G_{l,c}[\cdot]=F_{l,c}[\cdot]F_{l,c}[\cdot]^{T}
\end{equation}
$S_{l,c}[\cdot]$ is the segmentation mask in each layer $l$ with the channel c. $\lambda_{g}$ is the weight to configure layer preferences of global losses $\ell_{gs}$ which calculated between raw input image and features which was extracted by feature extraction network.$\lambda_{l}$ is the weight to configure layer preferences of local losses $\ell_{ls}$ which calculated between input segmentation image and features which was extracted by feature extraction network with the input of segmentation image.

We formulate the style transfer objective by combining both two components together:
\begin{equation}
L_{total}=\sum_{l=1}^L\alpha_l\ell_{feat}^l+\sum_{l=1}^L\beta_l\ell_{style}^l
\end{equation}
where L is the total number of convolutional layers and $l$ indicates the $l$-th convolutional layer of the deep convolutional neural network. $\alpha_{l}$ and $\beta_{l}$ are the weights to configure layer preferences. $\ell_{feat}$ is the content loss (Eq.(4)) and $\ell_{style}$ is the style loss(Eq.(9)). $\alpha_l$,$\beta_l$ are scalars, $\alpha_l=10^{2}$,$\beta_l=10^{4}$, in all cases the hyperparameters $\alpha_l$,$\beta_l$ are exactly the same.  We find that unconstrained optimization of Equation 18 typically results in images whose pixels fall outside the range [0,255]. For a more fair comparison with our method whose output is constrained to this range, for the baseline we minimize Equation 18 using projected L-BFGS. Image O is generated by solving the problem
 \begin{equation}
O=\arg\min_{I}{L_{total}}+\theta\ell_{TV}(I)
\end{equation}
where I is initialized with white noise. The advantage of this solution is that the requirement for mask is not too precise. It does not only retain the desired structural features, but also enhance the estimation of the pupil and iris information during the reconstruction of the style.
\begin{table}[!hbp]
\centering
\label{tabLr1}
\caption{Network architecture used for feature extraction network. }
\begin{tabular}{|l|l|}
\hline
Layer & Activation size\\
\hline
Input  &$3\times256\times256$ \\

Reflection Padding($24\times24$)&$3\times280\times280$ \\

$32@9\times9$ conv & $32\times280\times280$ \\

$64@3\times3$ conv & $64\times140\times140$ \\

$128@3\times3$ conv & $128\times70\times70$ \\

Residual block, 128 filters & $128\times66\times66$ \\

Residual block, 128 filters & $128\times62\times62$ \\

Residual block, 128 filters & $128\times58\times58$ \\

Residual block, 128 filters & $128\times54\times54$ \\

$64@4\times4$ deconv & $64\times128\times128$ \\

$32@4\times4$ deconv & $32\times256\times256$ \\

$3@9\times9$ deconv & $3\times256\times256$ \\
Output & $3\times256\times256$ \\
\hline
\end{tabular}
\end{table}

\section{Experimental Results}\label{sec:experimentalresults}

We experimented with two tasks: style transfer and appearance-based gaze estimation. Previous style style transfer work has used optimization to generate images; our feed-forward structure gives similar qualitative results, but the speed is increased by three orders of magnitude. Previous work on appearance-based gaze estimation has used fine synthetic images for training and real images for testing, or training with real images and testing with fine synthetic images. By using simulated data or purified real data for training, and using purified real data for testing, we can get encouraging qualitative and quantitative results.

\subsection{Style Transfer}

The purpose of the style transfer is to generate an image that combines the content of the target content image as the real image content with the style of the target style image as the style of the synthetic image. We train an image transformation network for each of the several hand selection style goals and compare our results with the baseline methods of Gatys et al.\cite{Gatys2015} and Feifei Li et al.\cite{lifeifei2016}. As a baseline, we re-implemented the method of Gatys et al.\cite{Gatys2015} and Feifei Li et al.\cite{lifeifei2016}. In order to make a fairer comparison with our method whose output is constrained to [0, 255], for the baseline, we minimize the equation 3 and equation 4 by using the projected L-BFGS by cropping the image to the range [0, 255] at each iteration. In most cases, the optimization converges to satisfactory results in 500 iterations.
\begin{figure*}[!htbp]
\centering
 \begin{minipage}[]{1\textwidth}
    \centering
     \includegraphics[width = 1\textwidth,angle=0]{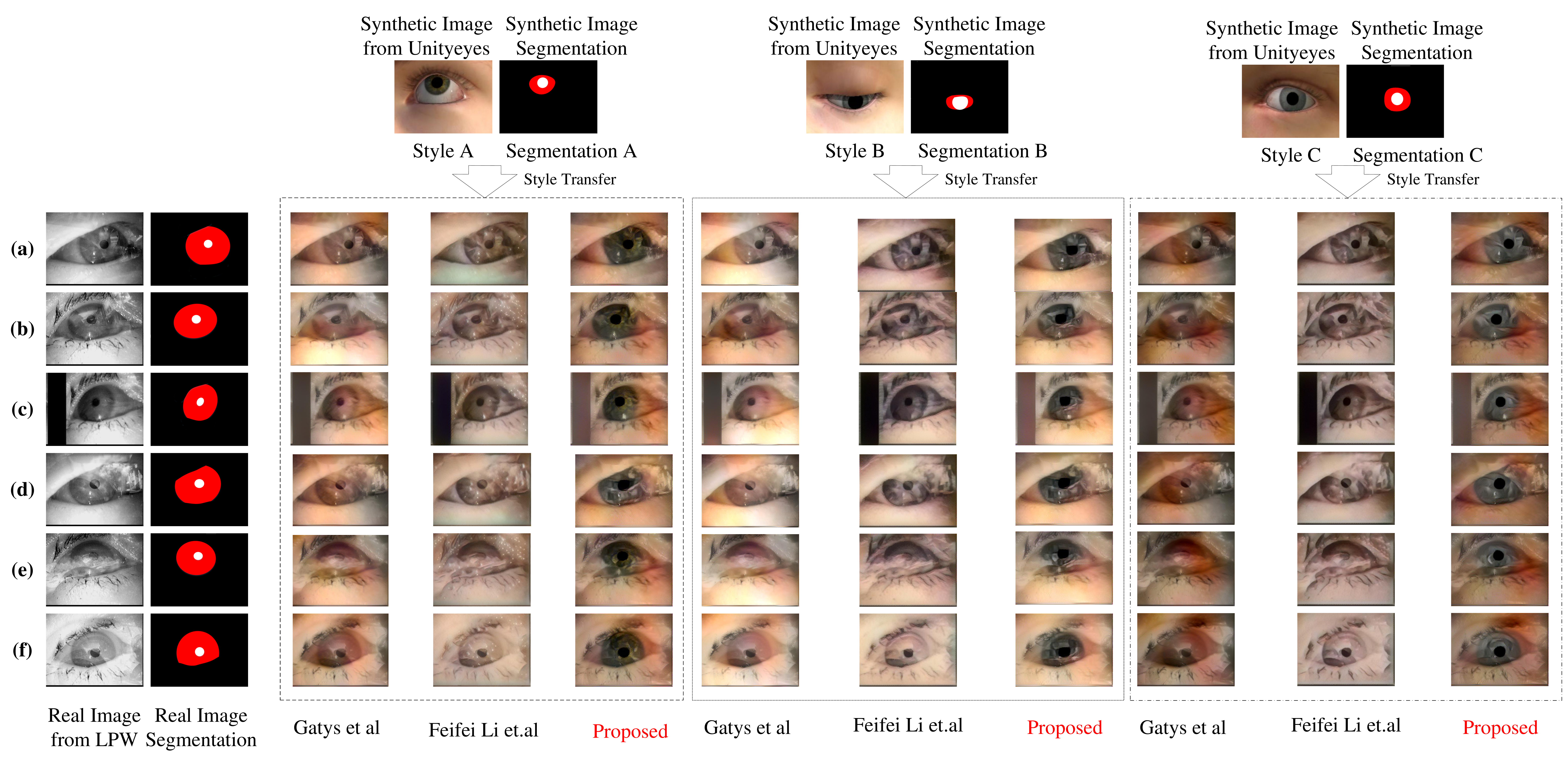}
 \end{minipage}
    \caption{Comparison on public LPW dataset with available style transfer methods.(a),(b),(c),(d),(e),and (f) represent the purified results of different distributions under six outdoor conditions from LPW dataset with three different styles from UnityEyes dataset. Style A, B, and C represent three different distributions of indoor conditions. The distribution of pupil and iris regions is dramatically different from style image. The proposed method, therefore can separate the pupil and the iris regions easily and the distribution of pupil and iris regions is similar to style image. }
    \label{figstyle}
\end{figure*}

$\mathbf{Implementation}$ $\mathbf{Details}$: We resize each of the 80 thousand training images to $256 \times 256$ and train our network with a batch size of 4 for 50000 iterations, giving roughly two epochs over the training data. We use Adam with a learning rate of $1 \times 10^{-4}$. The output images are regularized with total variation regularization with a strength of between $1 \times 10^{-7}$ and $1 \times 10^{-5}$. We choose conv$4\_2$  as the local content representation, and conv$1\_1$, conv$2\_1$, conv$3\_1$, conv$4\_1$ and conv$5\_1$ as the local style representation. conv$3\_2$  as the global content representation, and conv$1\_2$, conv$2\_2$, conv$3\_3$, conv$4\_3$ and conv$5\_3$ as the global style representation. Our implementation use Torch7 and cuDNN, training takes roughly 3 hours on a single GTX Titan X GPU.

\begin{figure*}[!htbp]
\centering
 \begin{minipage}[]{1\textwidth}
    \centering
     \includegraphics[width = 1\textwidth,angle=0]{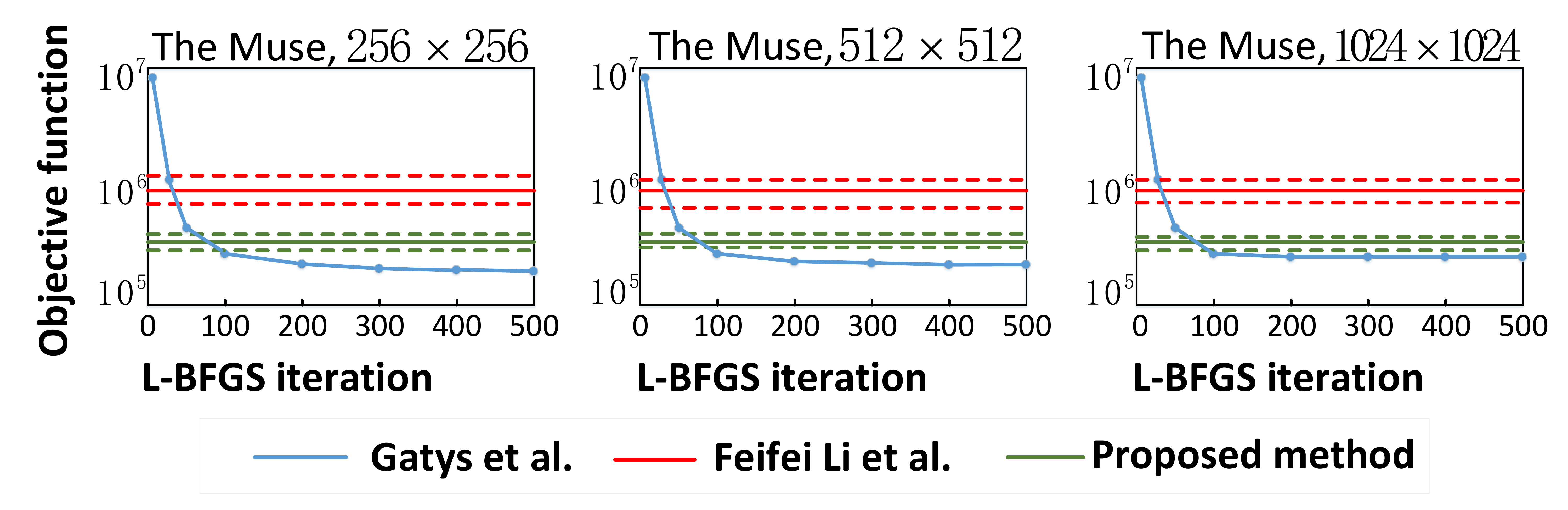}
 \end{minipage}
    \caption{Proposed style transfer networks and Gatys et al.\cite{Gatys2015} and Feifei Li et al.\cite{lifeifei2016} minimize the same objects. We compare their object values on 50 images; dashed lines and error bars show standard deviations. Our networks are trained on $256 \times 256$ images but generalize to larger images. }
    \label{figstyle}
\end{figure*}
$\mathbf{Qualitative}$ $\mathbf{Results}$: Fig.7 describes the style transfer method proposed in comparison to methods proposed by  Gatys et al.\cite{Gatys2015} and Feifei Li et al.\cite{lifeifei2016} across series of indoor and outdoor scenes from UnityEyes\cite{Wood2016learning} and LPW\cite{LPW2016} datasets respectively. (a),(b),(c),(d),(e), and (f) represents six different conditions of outdoor scenes from LPW dataset. On the other hand, styles A, B, and C from the UnityEyes dataset represent three different distributions of indoor conditions, which if closely observed, it can be seen that none of these styles has similar gaze angle with real images.

From (a),(b), and (c), it can be observed that the proposed method is less affected by light and achieves similar results with Gatys et al. \cite{Gatys2015} and Feifei Li et al.\cite{lifeifei2016}, but the proposed method can better preserve the color information of style image. From (d),(e), and (f), it can be seen that Gatys et al.\cite{Gatys2015} and Feifei Li et al.\cite{lifeifei2016} are influenced by light and other factors, the pupil and the iris cannot be completely separated. What's more, the distribution of pupil and iris regions is dramatically different from style image. The proposed method, therefore can separate the pupil and the iris regions easily and the distribution of pupil and iris regions is similar to style image.

Furthermore, it can be observed that no matter how style image changes, the distribution of the purified image is more inclined to that of the style image, which changes slightly according to different style images. However, the distribution of Gatys et al.\cite{Gatys2015}, Feifei Li et al.\cite{lifeifei2016} is more complex, because of light and other external factor interference, making it difficult to learn for gaze estimation tasks. Note that the proposed method preserves the annotation information while purifying the illumination of the real images.

$\mathbf{Quantitative }$ $\mathbf{Results}$: As evidenced by Gatys et al.\cite{Gatys2015} and Feifei Li et al.\cite{lifeifei2016} and reproduced in Figure 8, the image that produces the minimized pattern reconstruction loss preserves the style characteristics of the target image, but does not preserve its spatial structure. Reconstruction from higher layers transfers large-scale structures from the target image.
The baseline and our methods both minimize equation 19. The baseline performs explicit optimization over the output image, while our method is trained to find a solution for any content image in a single forward pass. We may therefore quantitatively compare the two methods by measuring the degree to which they successfully minimize Equation 19.

We used Pablo Picasso's Muse as a style image to run our method and baseline method on 50 images of the MS-COCO validation set. For the baseline method, we record the value of the objective function for each optimization iteration. For our method, we record Equation 19 for each image. From Figure 9, we can see that Feifei Li et al.\cite{lifeifei2016} achieved high losses, and our method achieved a loss comparable to 0 to 80 explicit optimization iterations.

Although our networks are trained to minimize Equation 19 for $256\times256$ images, they  are also successful in minimizing the objective when applied to larger images. We repeat the same quantitative evaluation for 50 images at $512\times512$ and $1024\times1024$, results are shown in figure 9. We can see that even at higher resolutions our method achieves a loss comparable to 50 to 100 iterations of the baseline method.

\begin{table}[!hbp]
\centering
\label{tabLr1}
\caption{ Speed (in seconds) for our style transfer networks vs Gatys et al. \cite{Gatys2015}, Feifei Li et al.\cite{lifeifei2016} for various resolutions. Across all image sizes, compared to 400 iterations of the baseline method, our method is three orders of magnitude faster than Gatys et al. \cite{Gatys2015} and we achieve better qualitative results (Fig.~\hyperref[figLPW]{7}) compared with Feifei Li et al.\cite{lifeifei2016} in tolerate speed. Our method processes $512\times512$ images at 20 FPS, making it feasible to run in real-time or on video. All benchmarks use a Titan X GPU.}
\begin{tabular}{|c|c|c|c|}
\hline
\diagbox{Method}{Time}{Resolution} & $256\times256$ & $512\times512$ & $1024\times1024$\\
\hline
Gatys et al. &12.69s & 45.88s & 171.55s\\
\hline
 Feifei li et al. & 0.023s & 0.08s & 0.35s \\
\hline
$\mathbf{Proposed Method}$ & $\mathbf{0.015s}$ &$\mathbf{0.05s}$ & $\mathbf{0.21s}$ \\
\hline
speedup (proposed vs Gatys) & $\mathbf{1060x}$ & $\mathbf{1026x}$ & $\mathbf{1042x}$ \\
\hline
speedup (proposed vs Feifei Li) & $\mathbf{1.53x}$ & $\mathbf{1.6x}$ & $\mathbf{1.67x}$ \\
\hline
\end{tabular}
\end{table}

$\mathbf{Speed}$: Table 2 compares the runtime of our method and Gatys et al.\cite{Gatys2015}, Feifei Li et al.\cite{lifeifei2016} for several image sizes. Across all image sizes, compared to 400 iterations of the baseline method, our method is three orders of magnitude faster than Gatys et al.\cite{Gatys2015} and we achieve better qualitative results (Fig.~\hyperref[figLPW]{7}) compared with Feifei Li et al.\cite{lifeifei2016} in tolerate speed. Our method processes images of size $512\times512$  at 20 FPS, making it feasible to run in real-time or on video.

\subsection{Appearance-based Gaze Estimation}

We evaluate our method for appearance-based gaze estimation on the MPIIGaze and purified MPIIGaze with base-line methods.
\begin{figure*}[!htbp]
\centering
 \begin{minipage}[]{1\textwidth}
    \centering
     \includegraphics[width = 1\textwidth,angle=0]{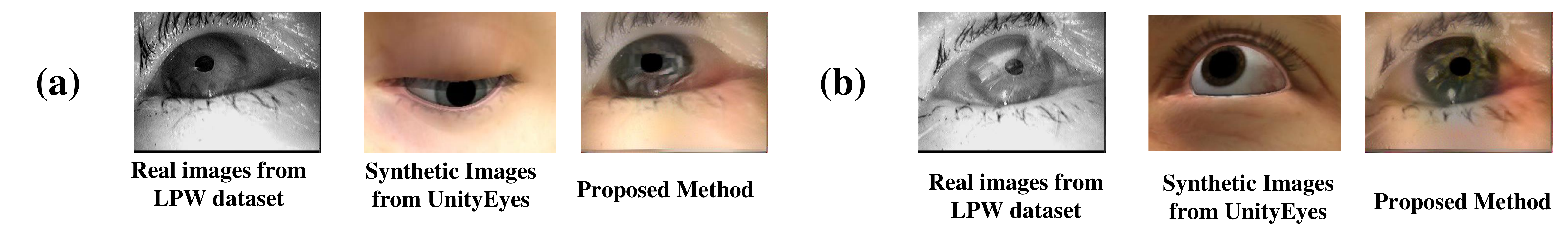}
 \end{minipage}
    \caption{Example output of proposed method for the LPW gaze estimation dataset. The skin texture and the iris region in the purified real images are qualitatively significantly more similar to the synthetic images than to the real images. }
    \label{figstyle}
\end{figure*}

$\mathbf{Implementation }$ $\mathbf{Details}$: In order to verify the effectiveness of the proposed method for gaze estimation, 3 public datasets (UTView\cite{Sugano2014learning}, SynthesEyes\cite{Wood2015rendering}, UnityEyes\cite{Wood2016learning}) are used to train the estimator with k-NN\cite{wang2017}, MPIIGaze dataset\cite{MPIIGaze2017} and purified MPIIGaze dataset (purified by proposed method) are used for test the accuracy. The eye gaze estimation network is similar to \cite{IJCNN2018}\cite{PR2018}, the input is a $35 \times 55$ gray scale image that is passed through 5 convolutional layers followed by 3 fully connected layers, the last one encoding the 3-dimensional gaze vector: (1)Conv 32$@$3$\times$3 (2)Conv 32$@$3$\times$3 (3)Conv 64$@$3$\times$3 (4)Max-Pooling 3$\times$3 (5)Conv 80$@$3$\times$3 (6)Conv 192$@$3$\times$3 (7)Max-Pooling 2$\times$2 (8)FC9600 (9)FC1000 (10)FC3 (11)Euclidean loss. All networks are trained with a constant $1\time10^{-3}$ learning rate and 512 batch size, until the validation error converges.

$\mathbf{Qualitative }$ $\mathbf{Results}$: Fig.9 shows examples of real, synthetic and purified real images from the eye gaze dataset. As shown, we observe a significant qualitative improvement of real images: Proposed method successfully captures the skin texture, sensor noise and the appearance of the iris region in the synthetic images. Note that our method preserves the annotation information(gaze direction) while purifying the illumination.

\begin{table}[!hbp]
\centering
\label{tabLr1}
\caption{Test performance on MPIIGaze and purified MPIIGaze; Purified MPIIGaze is the dataset which purified by proposed method. "Method" represents training set used with gaze estimation method.  Note how purifying real dataset for training lead to improved performance. }
\begin{tabular}{|c|c|c|}
\hline
Method & MPIIGaze & purified MPIIGaze \\
\hline
Support Vector Regression(SVR) &16.5$^{\circ}$ & 14.3$^{\circ}$ \\
\hline
Adaptive Linear Regression(ALR) &16.4$^{\circ}$ & 13.9$^{\circ}$ \\
\hline
Random Forest(RF) & 15.4$^{\circ}$ & 14.2$^{\circ}$ \\
\hline
KNN with UTview &16.2$^{\circ}$ & 13.6$^{\circ}$ \\
\hline
CNN with UTview &13.9$^{\circ}$ & 11.7$^{\circ}$ \\
\hline
KNN with UnityEyes  &12.5$^{\circ}$ & 9.9$^{\circ}$ \\
\hline
CNN with UnityEyes &9.9$^{\circ}$ & 7.8$^{\circ}$ \\
\hline
KNN with Syntheyes  &11.4$^{\circ}$ &8.0$^{\circ}$  \\
\hline
CNN with Syntheyes  &13.5$^{\circ}$ & 8.8$^{\circ}$ \\
\hline
\end{tabular}
\end{table}
\begin{figure*}[!htbp]
\centering
 \begin{minipage}[]{1\textwidth}
    \centering
     \includegraphics[width = 0.8\textwidth,angle=0]{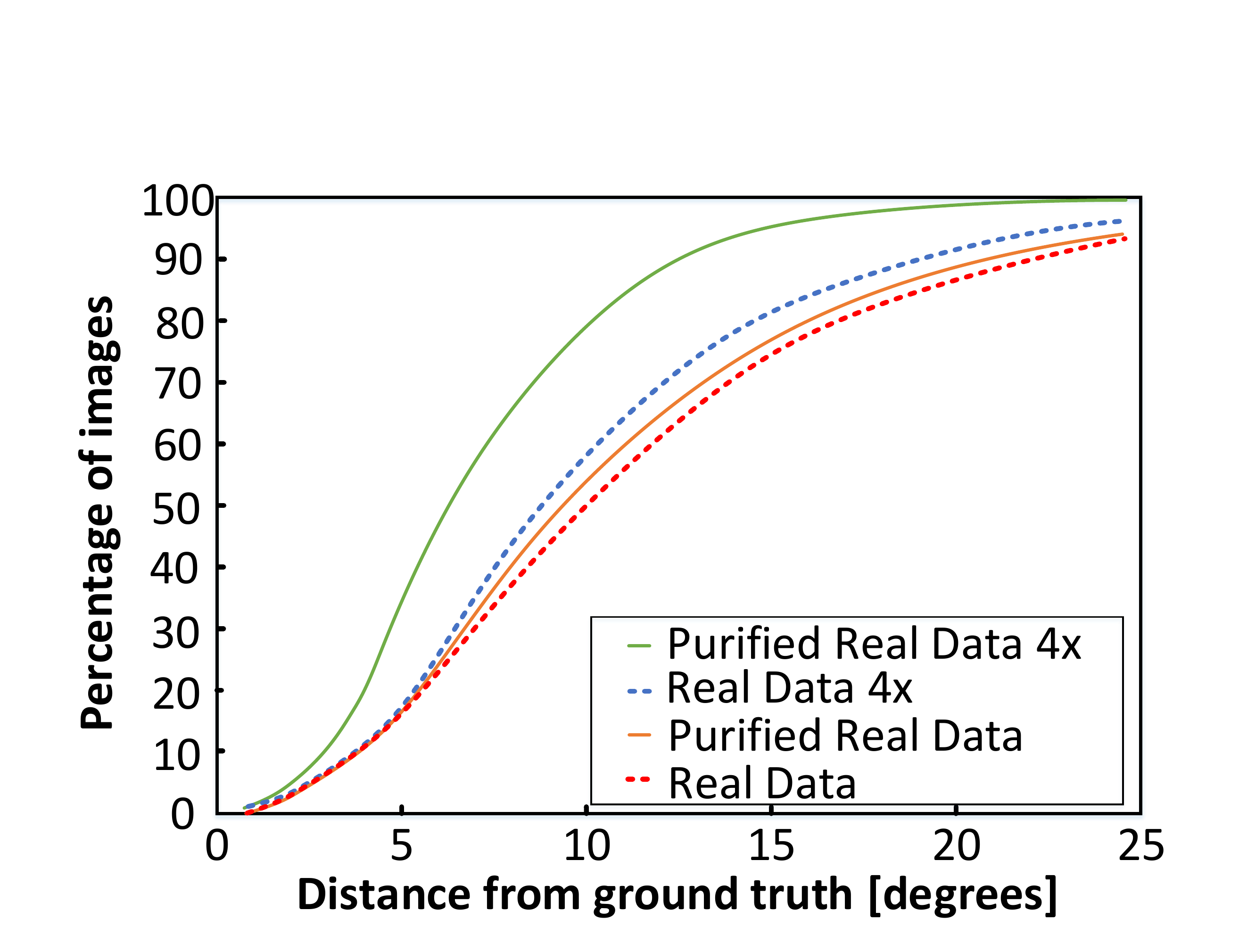}
 \end{minipage}
    \caption{Quantitative results for appearance-based gaze estimation on the MPIIGaze dataset and purified MPIIGaze dataset. The plot shows cumulative curves as a function of degree error as compare to the ground truth eye gaze direction, for different numbers of testing examples of data.}
    \label{figstyle}
\end{figure*}
$\mathbf{Quantitative }$ $\mathbf{Results}$: Five gaze estimation methods are used as base-line estimation methods. In addition to common methods such as Support Vector Regression(SVR), Adaptive Linear Regression(ALR) and Random Forest(RF), two methods are reproduced for fairly comparison with state-of-the-art. First method is a simple cascaded method\cite{wang2017}\cite{ACML2018}\cite{ICIMCS2017} which uses multiple $k$-NN($k$-Nearest Neighbor) classifier to select neighbors in feature space joint head pose,pupil center and eye appearance. The other method is to train a simple convolutional neural network (CNN)\cite{wild2017}\cite{IJCNN2018}\cite{PR2018} to predict the eye gaze direction with $ l_{2}$ loss. We train on UnityEyes ,UTView, SynthesEyes and test on MPIIGaze, purified MPIIGaze which is purified by proposed method. When testing on the MPIIGaze dataset, the training data can be either a raw dataset or a synthetic dataset, and when tested on a purified MPIIGaze dataset, the training data is either a purified real dataset or a raw synthetic dataset. Table 3 compares the performance of these two gaze estimation methods with different datasets. "Method" represents training set used with gaze estimation method. Large improvement in performance of testing on the output of proposed method is observed, each dataset improves at least three degrees of gaze estimation accuracy. This improvement shows the practical value of our method in many HCI tasks.

$\mathbf{Preserving }$ $\mathbf{Ground}$ $\mathbf{Truth}$: To quantify that the ground truth gaze direction doesn't change significantly, we manually labeled the ground truth pupil centers in 200 real and purified images by fitting an ellipse to the pupil. This is an approximation of the gaze direction, which is difficult for humans to label accurately. The absolute difference between the estimated pupil center of real and corresponding purified images is quite small: 0.8 $\pm$ 1.1 (eye width=55px)

\section{Conclusion}\label{sec:conclusion}

This paper took the first step to purify the real image by weakening its distribution, which is a better choice than improving the realism of synthetic image. We have applied this method to style transfer and gaze estimation tasks where we achieved comparable performance and drastically improved speed compared to existing methods. Performance evaluation indicates that purified MPIIGaze dataset (purified by our proposed method) recorded smaller error angle when used for gaze estimation task as compared with the raw MPIIGaze dataset.

In future, we intend to explore modeling the real-time gaze estimation system based on the proposed method and improve the speed of purifying videos.





\section*{ACKNOWLEDGMENTS}

The authors sincerely thank the editors and anonymous reviewers for the very helpful and kind comments to assist in improving the presentation of our paper. This work was supported in part by the National Natural   Science Foundation of China Grant 61370142 and Grant 61802043, by the   Fundamental Research Funds for the Central Universities Grant 3132016352,   by the Fundamental Research of Ministry of Transport of P. R. China Grant   2015329225300, by the Dalian  Science and Technology Innovation Fund 2018J12GX037 and Dalian Leading talent Grant, by the Foundation of Liaoning Key Research and Development Program.

\end{document}